\documentclass[letterpaper, 10 pt, conference]{ieeeconf}  

\IEEEoverridecommandlockouts                              

\usepackage{graphics} 
\usepackage{graphicx}
\usepackage{booktabs}
\usepackage{amsmath,amssymb,amsfonts}
\usepackage{textcomp}
\usepackage{xcolor}
\usepackage{subcaption}
\usepackage{amsmath} 
\DeclareMathOperator*{\argmax}{argmax}
\usepackage{amssymb}  
\usepackage{algorithm}
\usepackage[noend]{algpseudocode}
\usepackage{lipsum}
\usepackage{gensymb}

\usepackage{amsthm}
\usepackage{float}
\let\labelindent\relax
\usepackage[inline]{enumitem}
\usepackage[breaklinks,colorlinks]{hyperref}

\newcommand{\rev}[1]{{#1}}

\title{\LARGE \bf
IOSG: Image-driven Object Searching and Grasping}

\author{Houjian Yu$^{1}$, Xibai Lou$^{1}$, Yang Yang$^{2}$ and Changhyun Choi$^{1}$\\
\thanks{* This work was supported in part by the Sony Research Award Program and NSF Award 2143730.}
\thanks{$^{1}$ The authors are with the Department of Electrical and Computer Engineering, Univ. of Minnesota, Minneapolis, USA
        {\tt\small \{yu000487, lou00015, cchoi\}@umn.edu}}
\thanks{$^{2}$ Yang Yang is with the Department of Computer Science and Engineering, Univ. of Minnesota, Minneapolis, USA
        {\tt\small yang5276@umn.edu}}
}

\begin{document}

\maketitle
\thispagestyle{empty}
\pagestyle{empty}

\begin{abstract}
When robots retrieve specific objects from cluttered scenes, such as home and warehouse environments, the target objects are often partially occluded or completely hidden. Robots are thus required to search, identify a target object, and successfully grasp it. Preceding works have relied on pre-trained object recognition or segmentation models to find the target object. However, such methods require laborious manual annotations to train the models and even fail to find novel target objects. In this paper, we propose an Image-driven Object Searching and Grasping (IOSG) approach where a robot is provided with the reference image of a novel target object and tasked to find and retrieve it. We design a Target Similarity Network that generates a probability map to infer the location of the novel target. IOSG learns a hierarchical policy; the high-level policy predicts the subtask type, whereas the low-level policies, explorer and coordinator, generate effective push and grasp actions. The explorer is responsible for searching the target object when it is hidden or occluded by other objects. Once the target object is found, the coordinator conducts target-oriented pushing and grasping to retrieve the target from the clutter. The proposed pipeline is trained with full self-supervision in simulation and applied to a real environment. Our model achieves a 96.0\% and 94.5\% task success rate on coordination and exploration tasks in simulation respectively, and 85.0\% success rate on a real robot for the search-and-grasp task. Please refer to our project page for more information: \url{https://z.umn.edu/iosg}.

\end{abstract}

\section{Introduction}

Grasping an occluded target object is challenging. Imagine that an assistive robot is given the reference image of a pen and tasked to search and fetch it from a drawer full of office supplies. The target pen might not be visible to the robot at the beginning or there is not enough space to execute the grasping action. It is also possible that multiple pens lie together in the drawer. The robot needs to remove the objects that cover the pens and perceive the target pen, and eventually fetch it.

\begin{figure}[t]
\includegraphics[width=\linewidth]{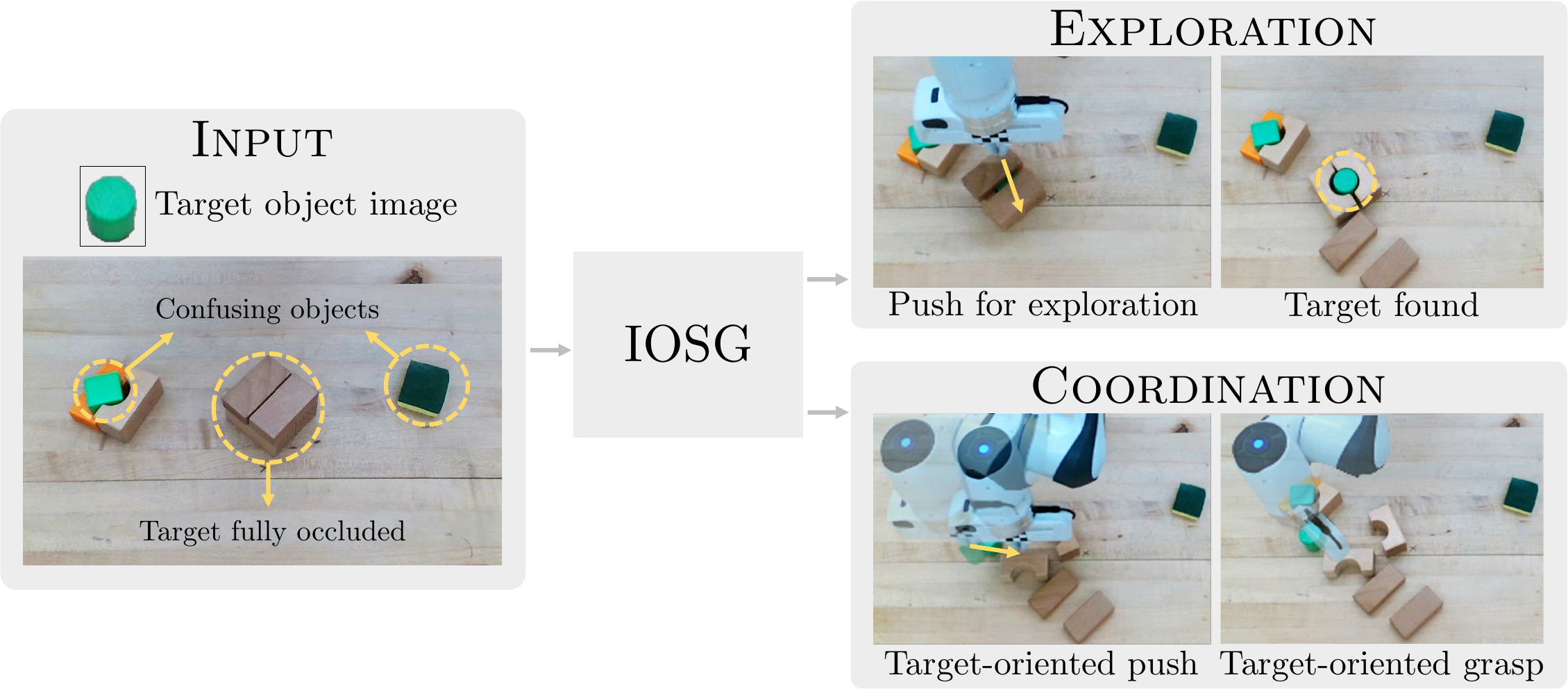}
\vspace{-5mm}
\caption{\textbf{Image-driven Object Searching and Grasping (IOSG).} The goal of IOSG is to find the target object queried by a reference image and to successfully grasp it. IOSG is challenging as the target object can be partially or fully occluded by other objects and there could be multiple confusing objects having similar shapes or colors to the target object, perplexing perception and decision-making processes.}
\label{front::fig}
\vspace{-5mm}
\end{figure}

As shown in Fig. \ref{front::fig}, to grasp an initially fully occluded target, the robot agent has to explore the workspace with an end-effector to disclose the hidden target object and then grasp it despite the clutter. Such a task has several challenges: 1) How to execute an efficient exploration to find the target?, 2) How to robustly reason about the target object if there are multiple similar objects that are never seen before?, 3) How to determine if the target is exposed enough and if grasping will be feasible?, and 4) How to learn a synergy between pushing and grasping? 

Yang et al.~\cite{yang2020deep} proposed Grasp The Invisible (GTI) and Danielczuk et al.~\cite{danielczuk2019mechanical} presented Mechanical Search to address some of the challenges. Although \cite{yang2020deep} manages to address the last challenge in coordination, the method relies highly on a pre-trained segmentation model determining if the target object is exposed and switches from exploration to coordination. \rev{Hence, the target object is limited to the object classes used in the segmentation model.} Such a method will fail when encountering novel objects or when there are confusing objects around. On the other hand, \cite{danielczuk2019mechanical} uses a heuristic method with a hard-coded threshold to maintain a priority list to select actions, and it mostly uses grasping actions to remove the clutter, limiting the synergy between pushing and grasping.

Most recent target-oriented grasping systems in robot manipulation focus on solving the target-driven grasping task that is equivalent to the GTI coordination subtask \cite{yang2020deep}, assuming that the target objects are visible without the necessity for searching or exploration. They usually require a reliable perception pipeline to recognize the target object from a dense clutter and assume manipulation capabilities to generate feasible grasping poses \cite{9197413,lou2021collision} or combining push and grasp motions \cite{huang2021visual,xu2021efficient}. 
However, when tasked with novel target objects, their perception is not reliable and hence the grasping performance degrades significantly. 

To address the aforementioned challenges, we propose Image-driven Object Searching and Grasping (IOSG) approach. \rev{Instead of training a segmentation model that is restricted to specific target labels,} our approach utilizes instance segmentation and a one-shot perception module to reason about the possible location of the target object. The high-level policy decides which low-level policies to execute depending on the target matching result, whereas the low-level policies (explorer and coordinator) execute the pushing and grasping actions accordingly. 

The main contributions of this work are as follows:
\begin{enumerate*}
   \item We employ a one-shot learning module, Target Similarity Network (TSN), to generate a target similarity map. Given the similarity map, the high-level policy makes an informed decision. 
   \item We propose a hierarchical search and grasp framework where the high-level policy determines the proper subtask type and chooses the corresponding low-level policy. 
   \item We present improved versions of the explorer and coordinator that outperform prior works in target object searching and grasping. 
\end{enumerate*}

\section{Related Works}

\subsection{Pushing and Grasping}

Synergistic object-agnostic pushing and grasping have been well explored \cite{zeng2018learning, huang2021dipn, deng2019deep, chen2020combining, yu2022self}. Zeng et al. \cite{zeng2018learning} propose the visual pushing for grasping (VPG) framework that jointly learns the pushing and grasping policies using a deep Q-network~\cite{Mnih15nature}, where the non-prehensile pushing facilitates the future object-agnostic grasping by creating more free space around objects. Yu et al. \cite{yu2022self} presented a singulation-and-grasping pipeline that first separates objects from each other and then conducts top-down agnostic grasping. Different from these target-agnostic methods, we aim to solve the target-oriented searching and grasping problem using a self-supervised approach. 

As for target-oriented robot manipulation methods \cite{yang2020deep, liu2022ge, xu2021efficient, 9880518, kurenkov2020visuomotor}, Yang et al. \cite{yang2020deep} introduce an effective object searching and grasping pipeline, which uses a Bayesian-based exploration policy to find the occluded target object and another classifier-based coordination policy to isolate the target object from the clutter and eventually grasps the target. Xu et al. \cite{xu2021efficient} propose a hierarchical framework that uses a deterministic grasping Q value as a threshold to decide the motion type. Liu et al. \cite{liu2022ge} train a sampling-based generator-evaluator architecture to remove the partially occluded target. Most of the aforementioned approaches use pre-trained foreground segmentation methods to track the target object, and they assume the segmentation results are with high accuracy. \rev{However, when encountering novel objects that have never been seen by the segmentation model, the erroneous target mask will negatively affect the performance of the policies. Our approach does not assume perfect segmentation masks and uses a similarity probability map to improve the generalization capability.} 

Though the object segmentation task has been extensively studied and a large number of generic object segmentation baselines are proposed, the laborious data collection process limits the pre-trained model usage in unstructured environments. Different from the target-driven object manipulation approaches where the policy network directly takes the target object segmentation masks or point clouds to infer desirable actions \cite{yang2020deep, liu2022ge, xu2021efficient, lou2021collision}, we use the object similarity image obtained from our target similarity network akin to a Siamese network \cite{koch2015siamese} to improve the generalization ability. \rev{Using visual encoding to match the corresponding target without a pre-trained segmentation model is widely used \cite{zeng2020transporter, seita_bags_2021, james2022q, james2022coarse}. While both Transporter Network \cite{zeng2020transporter} and Goal-conditioned Transporter Network \cite{seita_bags_2021} learn task-specific pick-and-place motions from expert demonstrations, our proposed method utilizes reinforcement learning training from scratch without human supervision.}

\subsection{Few-Shot Learning}

Our perception module is related to the Siamese convolutional neural network \cite{koch2015siamese}, prototypical network \cite{snell2017prototypical}, and the relation network \cite{sung2018learning}. These models use nearest neighbor \cite{snell2017prototypical}, linear classifier \cite{koch2015siamese,matching}, or a learnable non-linear classifier \cite{sung2018learning} to calculate the similarity score of image pairs to predict class labels. Specifically, \cite{koch2015siamese} uses fixed metrics (normally L1 or L2 distance measurement) for the feature distance computation. The main differences between our proposed model and the aforementioned are: 1) We exploit the depth data in object perception, whereas the prior works only use RGB data for training and testing. 2) Unlike relation network \cite{sung2018learning} that uses only shallow convolution layers for feature embedding and reasoning, we utilize deep neural network architecture (e.g., ResNet \cite{he2016deep}) for sophisticated RGB-D data encoding. 

\section{Proposed Approach}

In this section, we propose the Image-driven Object Searching and Grasping (IOSG) approach that enables robot manipulators to explore fully occluded objects with one target object image and conduct target-oriented pushing and grasping.


\begin{figure*}[t]
\begin{center}
    \includegraphics[width=1.0\linewidth]{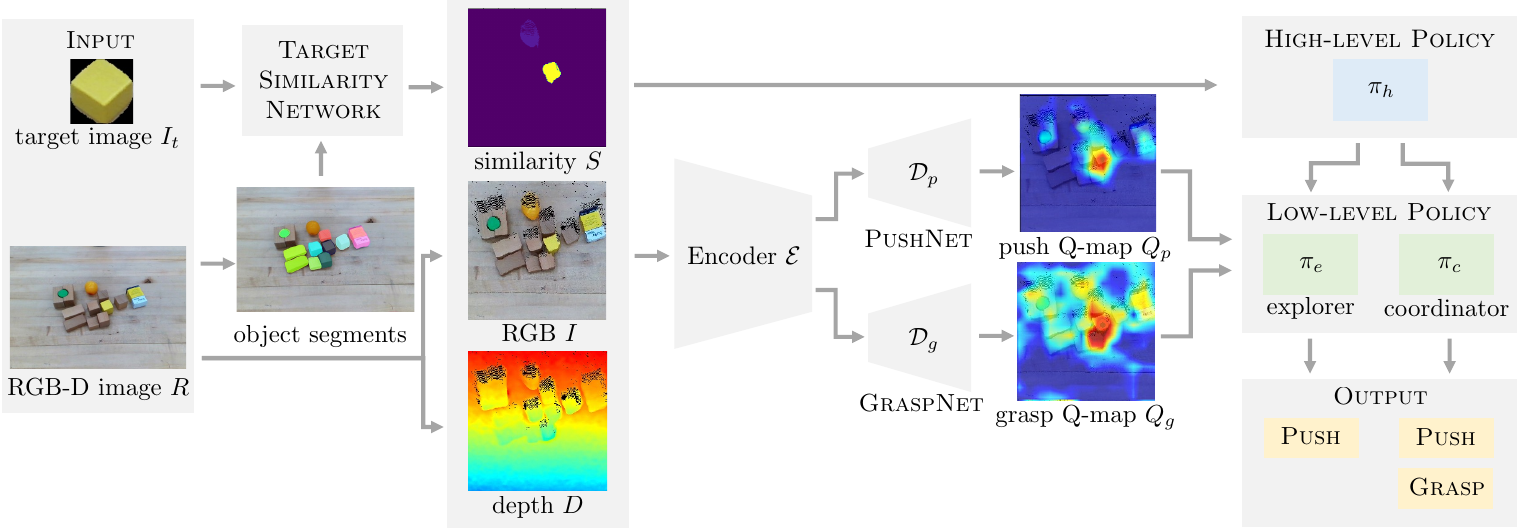}
\end{center}
\vspace{-3mm}
\caption{\textbf{Image-driven Object Searching and Grasping (IOSG) Pipeline.} The robot system is provided with a reference image $I_t$ of the novel target object. The target similarity network takes the object segments and the reference image $I_t$ as the input and outputs a similarity projection map $\mathcal{S}$. The high-level policy predicts the subtask type. The deep Q-Network encodes perception inputs and predicts push and grasp Q maps. Based on the prediction from the high-level policy, the low-level policies use the motion Q maps and domain knowledge to search for the target object or coordinate between push and grasp to remove the target from the clutter.}
\label{sys:fig}
\vspace{-3mm}
\end{figure*}


\subsection{System Overview}

Fig.~\ref{sys:fig} illustrates the overall pipeline of IOSG. 
The image of a target object $I_t$ and an RGB-D image $R$ are given as input to the pipeline. To match the target object and the objects in the RGB-D image, an object instance segmentation method first finds object segments from the RGB-D image. Target Similarity Network (TSN) then calculates the object-wise matching scores to obtain a target similarity map $S$ from the reference target image $I_t$ and the object segments (see the details of TSN in Section \ref{perception}). 

The RGB-D and target similarity maps are orthographically projected in the gravity direction to construct the RGB projection map $I \in \mathbb{R}^{H\times W\times 3}$, depth heightmap $D \in \mathbb{R}^{H\times W\times 1}$, and similarity projection map $S \in \mathbb{R}^{H\times W\times 1}$. 
The visual encoder $\mathcal{E}$ converts the visual input 
($S$, $I$, $D$) 
to a lower dimensional tensor, from which PushNet and GraspNet decoders reconstruct to push and grasp Q-maps ($Q_p$ and $Q_g$), respectively. Although not visualized in Fig.~\ref{sys:fig} for brevity, the input images are rotated with multiple angles to estimate the Q-maps with different orientations, allowing robots to reason an optimal pushing direction and grasping orientation~\cite{yang2020deep}.

Given the target similarity map $S$, the high-level policy $\pi_h$ predicts the current subtask type and chooses either exploration policy (explorer) $\pi_{e}$ or coordination policy (coordinator) $\pi_{c}$. As output, the actions (\textsc{Push}, \textsc{Grasp}) are determined from the low-level policies by taking into account the domain knowledge and the Q maps, $Q_p, Q_g$ (see details in Section \ref{high-level policy} and \ref{low-level policies}).

\subsection{Target Similarity Network} \label{perception}

\begin{figure}[!t]
\includegraphics[width=1.0\linewidth]{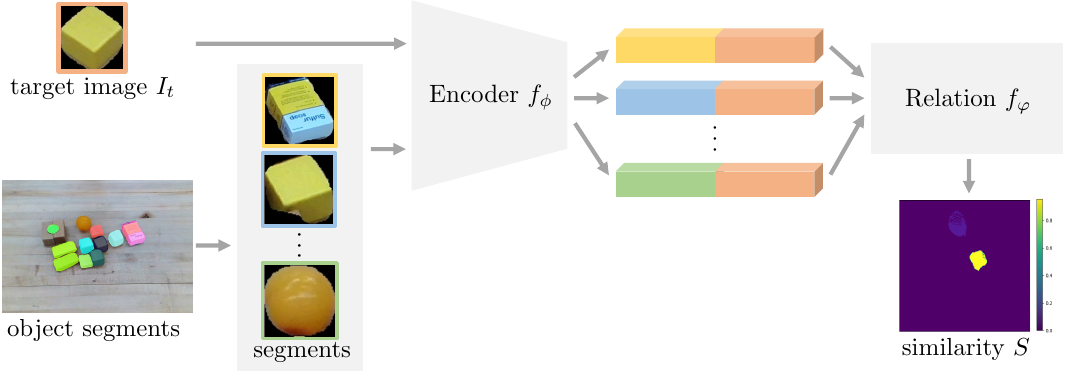}
\vspace{-5mm}
\caption{\textbf{Target Similarity Network (TSN).} TSN takes the object segments as input and generates the target similarity image ${S}$.}
\label{modules::fig}
\vspace{-2mm}
\end{figure}

Target Similarity Network (TSN) is inspired by few-shot learning for novel object matching \cite{koch2015siamese, sung2018learning} . TSN aims to recognize the target object from a reference image and output a probability map for target localization. 
As shown in Fig. \ref{modules::fig}, the visual encoder $f_{\phi}$ extracts visual features of the given target image and the object segments from the input RGB-D image. The visual feature of the target image is concatenated with each object segment feature. The relation network $f_{\varphi}$ calculates the similarity scores between the target image and the object segments and then fills the object segment masks with the similarity scores (i.e., object matching probabilities) to return the object similarity image $S$. 
A major difference between our TSN and the module in \cite{danielczuk2019mechanical} is that we utilize the whole similarity map to condition the downstream networks rather than using a heuristic threshold and only consider applying actions to the chosen object.

We follow the standard few-shot episode-based training in \cite{matching}. Unlike a traditional RGB-based Siamese network \cite{koch2015siamese}, we additionally consider the depth channel to differentiate the challenging objects (e.g., similar visual appearance but different in shape). We adopt a ResNet50 \cite{he2016deep} backbone as the encoder $f_{\phi}$ and two fully connected layers for similarity score prediction as a relation module $f_{\varphi}$ \cite{sung2018learning}. 


\subsection{Deep Q-Network}


We model the target-oriented pushing and grasping problem as a discrete Markov Decision Process (MDP). We use $s_t=(S,I,D) \in \mathbb{R}^{H\times W\times 5k}$ as the state representation, where we rotate the images $k$ times before forwarding in the network to reason about multiple orientations for motions. We set $k = 16$ with a fixed step size of $22.5^\circ$ w.r.t. the z-axis. 

The encoder $\mathcal{E}$ embeds $s_t$ and passes the feature to 
PushNet $\mathcal{D}_p$ and GraspNet $\mathcal{D}_g$ to predict the Q-maps in which each pixel value represents the expected future return if the corresponding motion is applied to the pixel location and the corresponding orientation. To maximize the reward, the pushing and grasping motion primitives are executed at the highest Q-value in the Q-maps~\cite{yang2020deep, zeng2018learning, Mnih15nature}.

\subsection{High-level Policy}
\label{high-level policy}
The previous work \cite{yang2020deep} uses a semantic segmentation model such that the robot agent performs the coordination subtask immediately when the target object segment is detected. However, when encountering novel target objects that have never been trained in the segmentation module, reliable detection of the targets is more difficult, and hence a proper determination of the subtasks becomes challenging.

To tackle this challenge, we relax the known-object assumption so that the robot can find a novel target object and take into account potentially noisy matching results. Specifically, we create a classifier-based high-level policy that takes as input the similarity image $S$ and determines the optimal low-level policy to execute. The high-level subtask classifier $f_{\pi_h}$ is formulated as

\begin{equation}
    \pi_h : \argmax_\pi \underset{\{\pi_e, \pi_c\}}{f_{\pi_h}}\!({S})
\end{equation}
where the classifier $f_{\pi_h}$ uses a deep ResNet18 backbone and outputs the type of the current subtask (i.e., exploration subtask or coordination subtask).

\subsection{Low-level Policies} \label{low-level policies}
\subsubsection{Explorer} Explorer aims to perform a sequence of pushing actions to unearth the hidden target object. The Bayesian-based exploration policy in \cite{yang2020deep} combines the push Q maps $Q_p$, the clutter prior $C_p$ (generated by detecting the varying heights), and the pushing failure map $F_p$. However, the exploration policy is not a universal solution for the exploration subtask. In \cite{yang2020deep}, the exploration policy is never trained and highly depends on the pushing Q maps which is learned from the other subtask, coordination. In our experiments, we observed that the Bayesian-based exploration policy cannot resolve the problem. To tackle the problem, we design an effective reward function (see Section~\ref{sec:reward}) and train the exploration policy through DQN training. During the exploration subtask training, we only apply push actions and update the DQN to encourage exploration actions.

\subsubsection{Coordinator} Coordinator determines a series of pushing and grasping actions to grasp the target object in clutter. To learn the push-and-grasp synergy in the coordination subtask, the classifier-based coordinator is trained to predict the action type (i.e., push or grasp) with domain knowledge features and visual inputs. The original coordination policy in \cite{yang2020deep} only considers five hand-coded features, including the maximum grasping Q value $q_g$, the maximum pushing Q value $q_p$, target border occupancy ratio $r_b$, target border occupancy norm $n_b$, and the number of consecutively failed grasps $c_g$.

\begin{figure}[h]
    \includegraphics[scale=0.4]{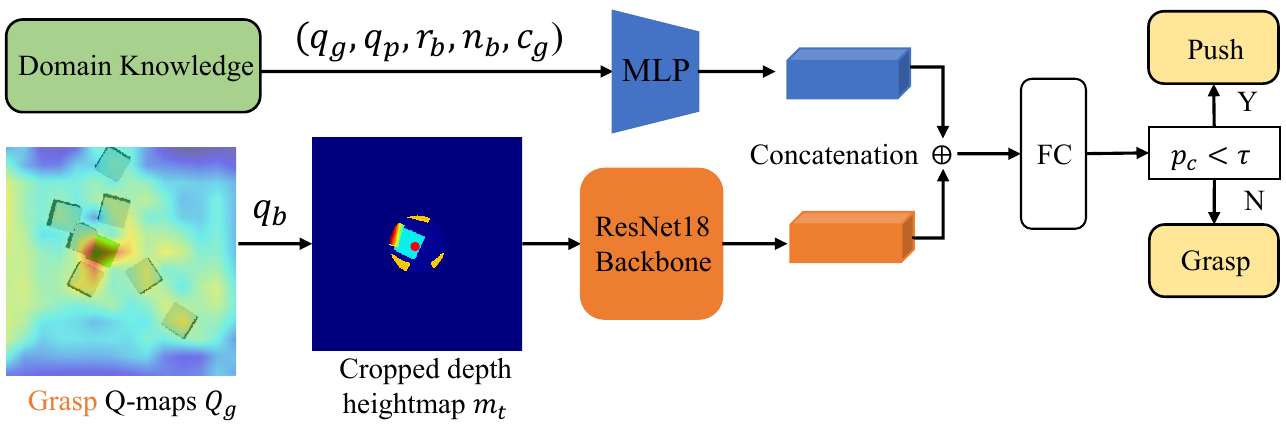}
    \vspace{-4mm}
   \caption{\textbf{\rev{Classifier $f_a$ in coordinator}}. The coordinator takes the domain knowledge features and the cropped depth heightmap $m_t$ as input and predicts the grasping confidence score $p_c$.}
   \label{coordinator::fig}
   \vspace{-2mm}
\end{figure}

Inspired by \cite{9880518}, we provide the coordinator with additional visual data: a cropped depth heightmap $m_t$ to predict the action type (push or grasp). In accordance with the open gripper's size, we preserve the depth heightmap $D$ values within 30 pixels from the best grasping pixel $q_b$ in grasp maps $Q_g$ and align the cropped area to the center of the image to derive $m_t$. The network details can be found in Fig. \ref{coordinator::fig}. The action classifier $f_a$ is formulated as
\begin{equation}
    p_c = {f_a}\! (q_g, q_p, r_b, n_b, c_g, m_t)
\end{equation}
If the grasping confidence score $p_c$ is lower than the threshold $\tau$, the robot agent will choose the push action, otherwise the grasp action. We use two fully connected layers with batch normalization (BN) \cite{ioffe2015batch} and ReLU \cite{nair2010rectified} to process the domain knowledge features and a ResNet18 to process the cropped depth heightmap $m_t$. We then concatenate the two processed features and finally pass two fully connected layers with BN and ReLU to predict the action type.
The overall Image-driven Object Searching and Grasping pipeline is summarized in Algorithm~\ref{alg::testing}.

\begin{algorithm}[t]
\caption{Image-driven Object Searching and Grasping}
\textbf{Input}: RGB-D image $R$ and reference image $I_t$\\
\textbf{Output}: motion primitives $a_t$ at time $t$
\begin{algorithmic}[1]
\State $\mathcal{M}$ $\gets$ \texttt{SegmentationModel}($R$)
\State ${S}$ $\gets$ \texttt{TargetSimilarityNetwork}($I_t$,$\mathcal{M}$)
\State $I, D$ $\gets$ \texttt{HeightmapProjection}($R$)
\State $s_t$ $\gets$ (${S}$, $I$, $D$)
\State $\pi_h$ $\gets$ $\argmax_\pi \underset{\{\pi_e, \pi_c\}}{f_{\pi_h}}\!({S})$ \Comment{high-level classifier}
\If{$\pi_h == \pi_e$} \Comment{exploration subtask}
\State $a_t$ $\gets$ $\underset{a}\argmax \, \mathcal{D}_p(s_t)$
\Comment{explorer}
\Else
\Comment{coordination subtask}
\State $Q_p \gets \mathcal{D}_p(s_t)$, $Q_g \gets \mathcal{D}_g(s_t)$
\State $p_c \gets {f_a}\! (q_g, q_p, r_b, n_b, c_g, m_t)$
\State $y \gets$ \texttt{Comparison}($p_c$,$\tau$)
\State $a_t \gets \underset{a}\argmax \, Q_y$
\Comment{coordinator}
\EndIf
\end{algorithmic}
\label{alg::testing}
\end{algorithm}
\vspace{-2mm}

\subsection{Reward Function}
\label{sec:reward}
\subsubsection{Explorer} We assign rewards to any actions that remove objects on top of the target object. When the target object is exposed slightly, we encourage the robot agent to push away the covering objects. The explorer reward function $R_e$ is defined below:
\begin{equation}
\begin{small}
    R_e=\begin{cases}0.5, &\text{area of $g_t$ increases by 50 pixels or} \\ &\text{push vector passes $g_t$}\\0.25, &\text{\# of objects above the target decreases} \\0, &\text{otherwise}
    \end{cases}
\end{small}
\end{equation}
where $g_t$ represents the target mask at time $t$.

\subsubsection{Coordinator} For pushing actions, we assign a reward of 0.25 if the pushing vector passes the actual target mask or if the target moves. To determine if the target has moved by a certain threshold, we count the number of pixels on the target mask that overlap before and after the pushing event. Moreover, if the border occupancy value $o_b$ (defined in \cite{yang2020deep}) decreases by some threshold, we assign the pushing action a 0.5 reward value. The pushing reward function $R_p$ is defined as:
\begin{equation}
\begin{small}
    R_p=\begin{cases}0.5, &\text{$o_b$ decreases by 0.1}\\0.25, &\text{push vector passes $g_t$ \& target moves} \\0, &\text{otherwise}
    \end{cases}
\end{small}
\end{equation}

Lastly, we define the grasping reward function $R_g$ as follows:
\begin{equation}
\begin{small}
    R_g=\begin{cases}1, &\text{target is successfully grasped} \\0.5, &\text{best grasp pixel $q_b$ is in target mask} \\0, &\text{otherwise}
    \end{cases}
\end{small}
\end{equation}

\section{Data Collection and Training Details} \label{training details}

The proposed system is trained by self-supervision in CoppeliaSim~\cite{6696520} simulation with Bullet \cite{coumans2015bullet} physics engine 2.83. 

\textbf{Target Similarity Network:} We collect the training/validation/testing data from the YCB \cite{7254318} dataset. Specifically, we generate the training set with 20 different YCB objects and 200 RGB-D images for each object category. We use a one-way one-shot manner for training. We combine the standard binary cross entropy (BCE) loss for the network outputs and the triplet loss~\cite{hermans2017defense} for the embedded features output from the ResNet50 backbone as the training loss.
We use the Adam \cite{kingma2014adam} optimizer with an initial learning rate of $5\times10^{-4}$. 

\begin{figure}[h]
    \centering
    \begin{subfigure}{0.238\linewidth}
        \includegraphics[width=\textwidth]{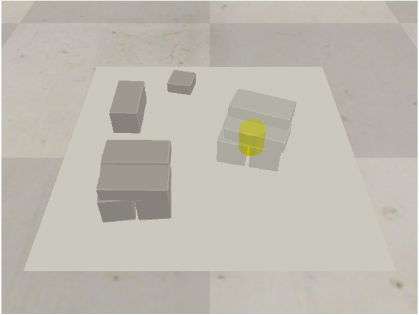}%
        \vspace{-5pt}
        \caption*{case 1}
    \end{subfigure}
    \begin{subfigure}{0.238\linewidth}
        \includegraphics[width=\textwidth]{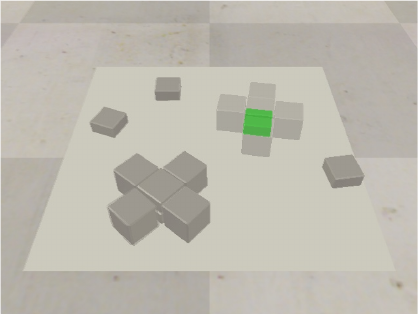}%
        \vspace{-5pt}
        \caption*{case 2}
    \end{subfigure}
    \begin{subfigure}{0.238\linewidth}
        \includegraphics[width=\textwidth]{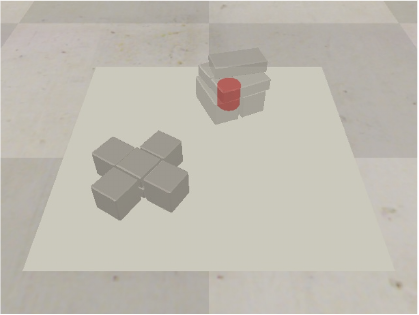}%
        \vspace{-5pt}
        \caption*{case 3}
    \end{subfigure}
    \begin{subfigure}{0.238\linewidth}
        \includegraphics[width=\textwidth]{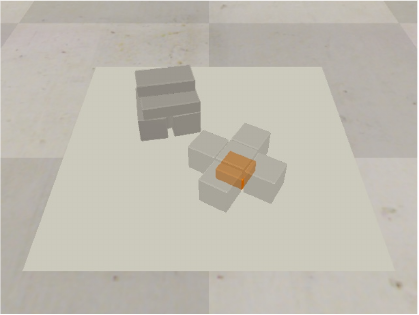}%
        \vspace{-5pt}
        \caption*{case 4}
    \end{subfigure}
    \vspace{-6mm}
    \caption{\textbf{Training cases for Explorer.} The target object in each case is initially invisible to the camera. There is only one target object (colored block) in each training case.}
    \vspace{-2mm}
    \label{fig:exploration-train}
\end{figure}

\textbf{Policies Training:} Multi-stage training is utilized. \rev{Following the object setting of \cite{yang2020deep}}, at the first 1000 iterations, we randomly drop $m=3$ obstacle objects and $n=7$ potential target objects into the workspace. The DQN is trained in an $\epsilon$-greedy policy. After 1000 iterations, we increase the number of obstacle objects to $m=8$. In the second stage from 2000 to 4000 iterations, the robot agent only executes grasping actions. We train the coordinator classifier $f_a$ and collect the coordinator training data (domain knowledge features as well as the cropped depth heightmap $m_t$) and the ground truth labels corresponding to grasping success or failure. In the third stage, from 4000 to 8000 iterations, we switch from ${\epsilon}$-greedy to coordinator $\pi_{c}$, where the DQN is jointly updated with the coordinator classifier. In the last stage, we train our DQN with the exploration subtask. We incorporate the exploration and coordination policies together on 4 predefined object searching and grasping cases as illustrated in Fig. \ref{fig:exploration-train}. \rev{Meanwhile, we collect the corresponding similarity projection maps ${S}$ to train the high-level subtask classifier}. We use the Huber loss to train the DQN, and BCE loss is used to train the coordinator classifier and the high-level subtask classifier.

\begin{figure}[h]
\centering
\includegraphics[scale=0.5]{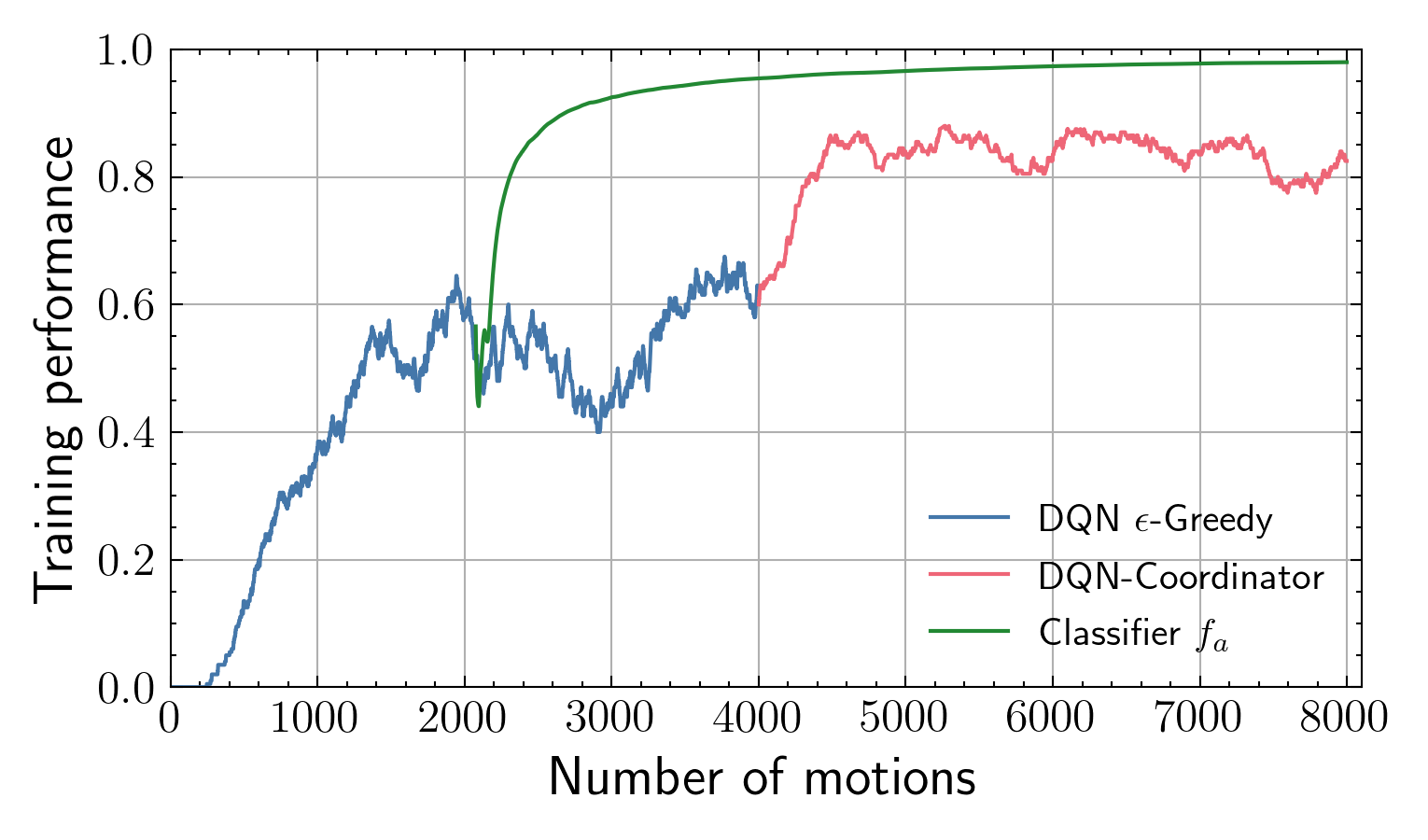}
\vspace{-1mm}
\caption{\textbf{Grasping success rate and coordinator prediction accuracy}. The blue curve and the red curve show the target grasping success rate in training. The green line indicates the prediction accuracy of the action classifier $f_a$ in coordinator $\pi_c$ in training.}
\label{curve}
\vspace{-2mm}
\end{figure}

Fig. \ref{curve} shows the training performance. We compute the latest 200 target-oriented grasping success rates through training iterations and the overall prediction accuracy of the coordinator classifier. The coordinator policy achieves a higher grasping success rate compared with the $\epsilon$-greedy policy.

\section{Experiments}

In this section, we conduct multiple experiments to evaluate the proposed approach. The experiment goals are to verify if: 1) our Target Similarity Network is able to identify the target object in the clutter with one reference image, 2) the proposed exploration policy performs well, and 3) IOSG outperforms other baselines in terms of both the task success rate and the motion efficiency. 

\subsection{Perception Experiments}

\begin{figure}[h]
  \centering
  \begin{subfigure}{0.27\textwidth}
    \includegraphics[width=\textwidth]{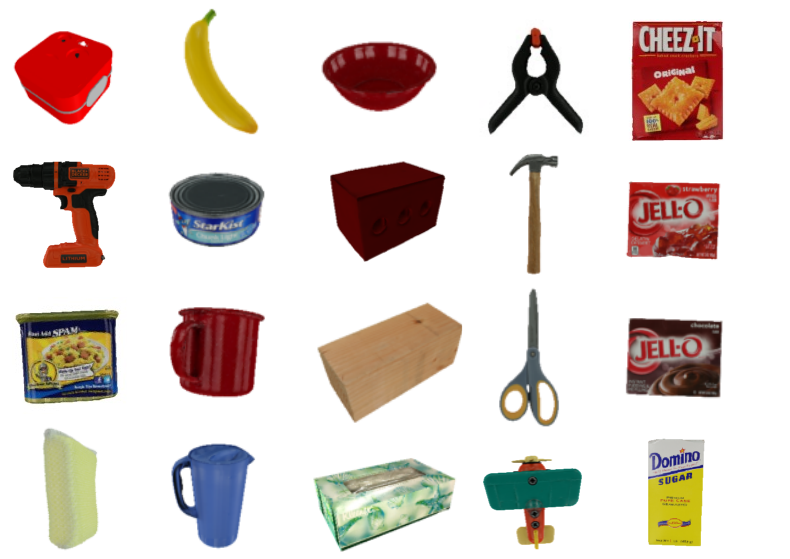}
    \caption{Simulated known}
    \label{fig:testing_sim}
  \end{subfigure}
  \begin{subfigure}{0.16\textwidth}
    \includegraphics[width=\textwidth]{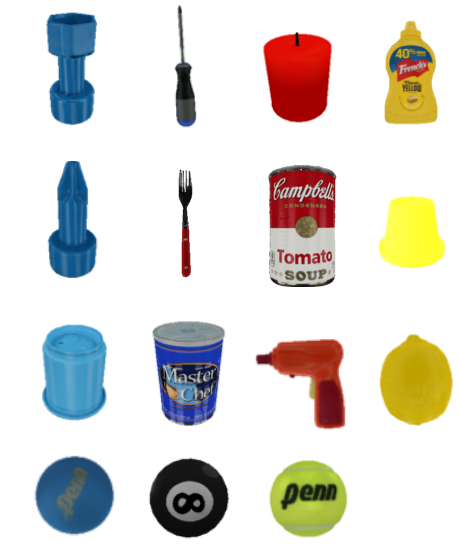}
    \caption{Simulated novel}
    \label{fig:testing_real}
  \end{subfigure}
  \caption{\textbf{Target Similarity Network training and testing objects.} We use 20 YCB objects to generate the train/validation/test dataset and prepare another test set with 14 novel objects.}
  \vspace{-2mm}
  \label{fig:testing}
\end{figure}

We test our Target Similarity Network (TSN) on known, novel, and simple simulated objects. Fig. \ref{fig:testing} shows the known and novel objects used in this experiment. 
The simple objects are pure-colored toy blocks (8 colors and 6 shapes) used during the policy training. Note that the novel objects have different shapes and textures from the training objects. We collected 1000, 500, and 1000 image pairs of known, novel, and simple simulated objects for comparison, respectively. 

We conduct an ablation study on TSN trained with different loss functions. We test our model with a 5-way 1-shot setting. Table \ref{tab2} summarizes the TSN performance tested with the known simulated objects. The experiment validates our loss function choice (BCE + Triplet).  
We also compare the matching performance with two baselines: one uses non-cropped object segments (i.e., each segment in the original input image size) with (RGB-D) and without (RGB) a depth channel. It shows the effectiveness of cropping segments and using depth channel. 

\begin{table}[h]
\caption{Target Similarity Network Accuracy (\%)}
\vspace{-3mm}
\label{tab2}
\begin{center}
\begin{tabular}{cccc}
\toprule
  & RGB & RGB-D & Crop RGB-D (Ours)\\
\midrule
BCE loss & 78.1 & 82.3  & 85.1\\
Triplet loss & \textbf{87.5} & 90.0  & 92.7\\
BCE $+$ Triplet & 86.9 & \textbf{90.3}  & \textbf{96.4}\\
\bottomrule
\end{tabular}
\vspace{-5mm}
\end{center}
\end{table}


\subsection{Exploration Subtask}

To validate the proposed exploration policy, we compare our approach with four baselines: 1) \textbf{GTI} \cite{yang2020deep} exploration policy. This exploration policy uses a Hadamard product of the push Q-map, the clutter prior map, and the Bayesian-based push failure map to predict the exploration push map. The clutter prior map is derived from the depth heightmap by detecting the difference between before and after shifting the depth heightmap by certain pixels. The push failure map is used to reduce the push probability around the most recent push failure area. 2) \textbf{IOSG-Coordinator} is a variant of our approach where we directly use the DQN to generate the push Q-map. The DQN is only trained in the first three stages for the coordination tasks. 3) \textbf{IOSG-Coordinator+Bayes} uses the same push Q-map as \textbf{IOSG-Coordinator} and is combined with the clutter prior map and the Bayesian-based push failure map. 4) \textbf{Ours+Bayes} uses the same push Q-map as our proposed exploration policy. The DQN is trained for the complete four stages. 

\begin{figure}[t]
\includegraphics[width=\linewidth]{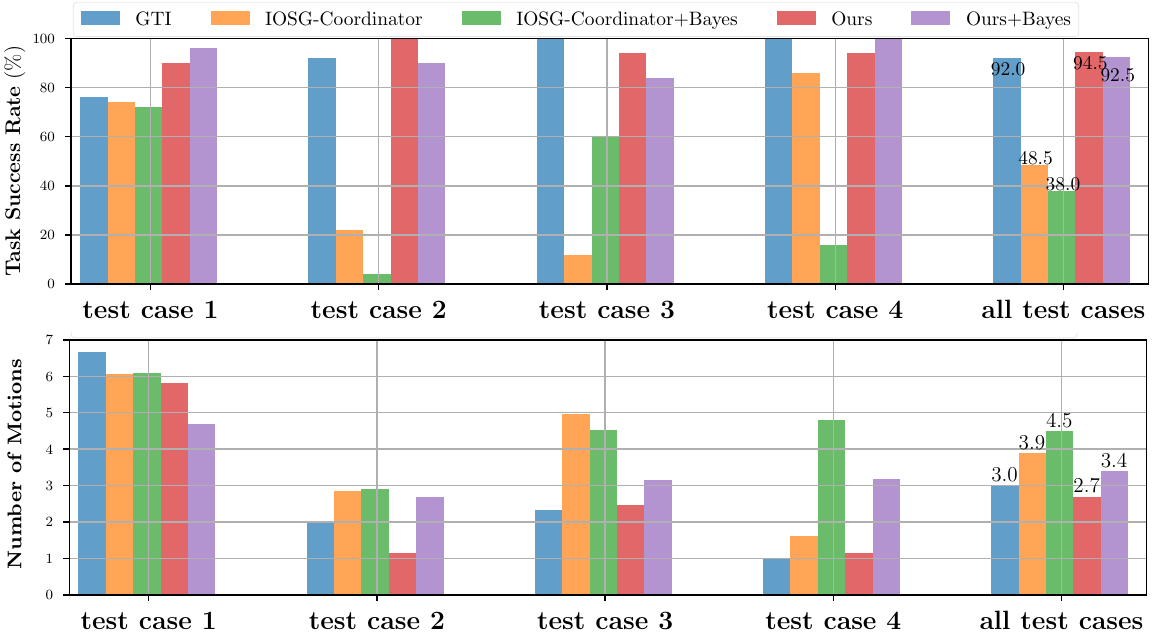}
\vspace{-4mm}
\caption{\textbf{Performance in exploration subtask.} 
Our proposed exploration policy achieves $94.5\%$ success rate with an average 2.7 motions}
\label{exploration::fig}
\vspace{-2mm}
\end{figure}

We evaluate the baseline methods with the same 4 exploration test cases from \textbf{GTI} \cite{yang2020deep}. A trial is considered to be successful if the target object is exposed by over 200 pixels. The maximum number of motions is $n_{motion} = 2 \times n_{cluster} - 1$, where $n_{cluster}$ represents the number of object clusters in the scene. We execute each test case 50 times and the results are shown in Fig. \ref{exploration::fig}.

Our proposed exploration policy achieves $94.5\%$ success rate with high motion efficiency compared with \textbf{GTI} baseline that achieves $92.0\%$ success rate. \textbf{IOSG-Coordinator} shows that our coordinator is not sufficient to complete the exploration subtask given that the coordinator has never been trained on the exploration subtask. \textbf{IOSG-Coordinator+Bayes} indicates that the Bayesian-based exploration policy is not an effective solution to the exploration subtask. Moreover, by comparing \textbf{IOSG-Coordinator} with \textbf{Ours}, it shows the effectiveness of our reward design. The task success rate increases significantly with such additional exploration policy training. \textbf{Ours+Bayes} shows slightly lower performance than \textbf{Ours}. The additional clutter prior map and the push failure map provide rough object cluster position. However, they cannot reason about the pushing orientation. In such a case, the robot is likely to push one single object rather than pushing toward the center of the cluster. Therefore, it is not able to complete the exploration subtask with high motion efficiency.

\subsection{Coordination Subtask}\label{Cooordination analysis}

\begin{figure}[t]
\includegraphics[width=\linewidth]{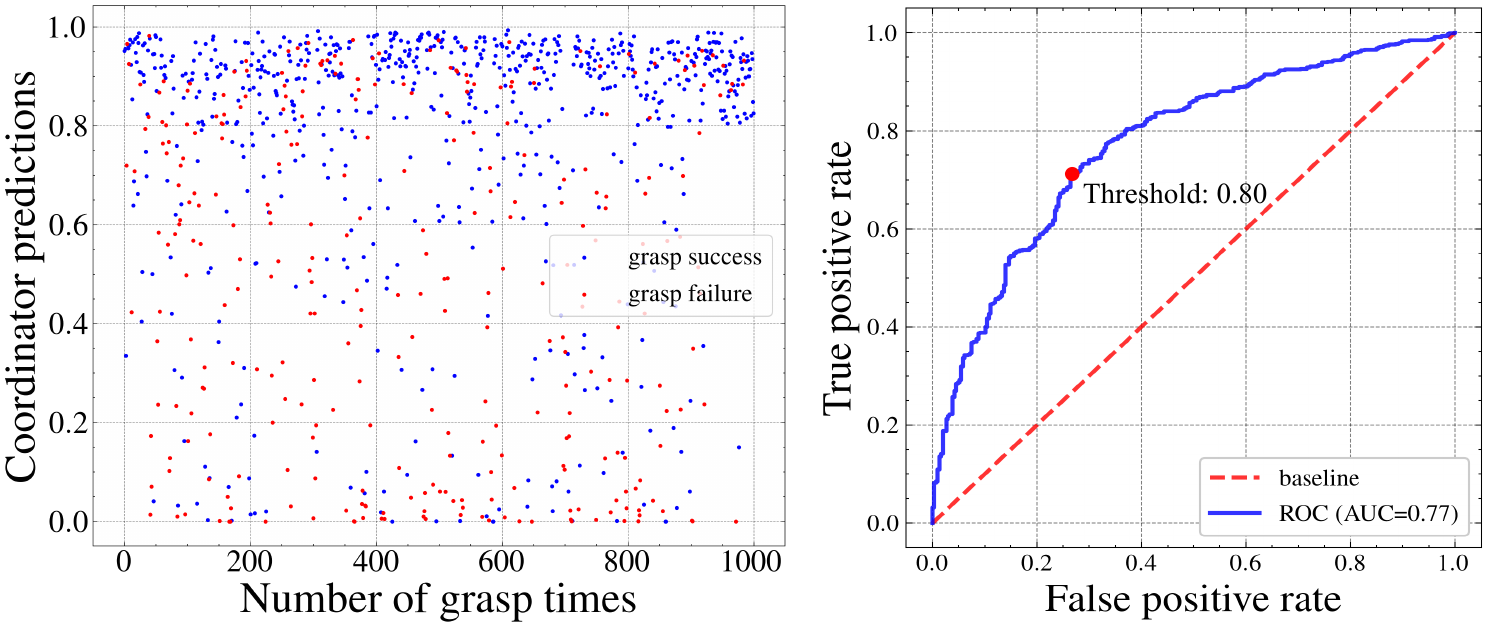}
\vspace{-6mm}
\caption{\textbf{Data visualization and ROC curve.} The left image shows the predicted grasping confidence score $p_c$ from the coordinator classifier $f_a$ and the grasping labels. The right image shows the ROC curve. The $AUC=0.77$ and the corresponding optimal threshold is $\tau=0.80$.}
\label{roc::fig}
\vspace{-2mm}
\end{figure}

\textbf{Parameter Verification:} We verify the threshold value $\tau=0.80$ used for coordinator classifier $f_a$ in Section \ref{low-level policies}. We collect the grasping confidence score $p_c$ predicted by the coordinator classifier $f_a$ and the grasping successes or \rev{failure} 1000 times. To determine the optimal grasping confidence threshold $\tau$ as well as evaluate the performance of the classifier, we use the standard AUC-ROC \cite{fawcett2006introduction} analysis. As shown in Fig. \ref{roc::fig}, the optimal threshold is $\tau=0.80$ such that the classifier has the best performance.

\begin{figure}[t]
\begin{center}
\includegraphics[width=\linewidth]{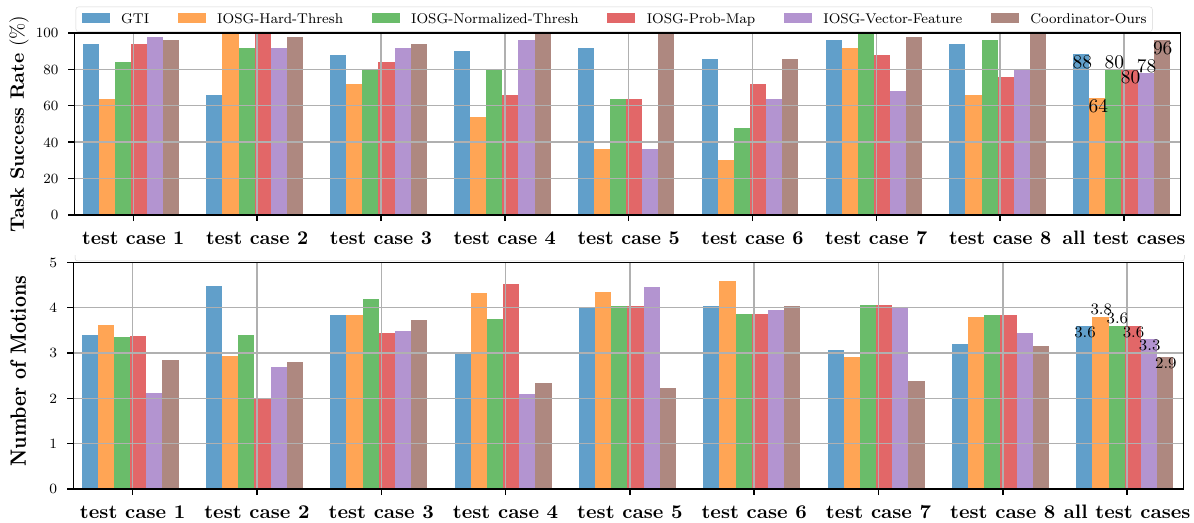}
\vspace{-6mm}
\end{center}
\caption{\textbf{\rev{Performance in coordination subtask}}. We show the task success rate and the average number of motions in 8 coordination tasks. Our proposed model achieves 96.0\% success rate with 2.9 motions. Please zoom in for more details.}
\vspace{-4mm}
\label{coordinator::performance}
\end{figure}

\textbf{Ablation Studies:} We compare the target grasping success rate as well as the average number of motions of the coordination subtask with several baselines. 1) \textbf{GTI} \cite{yang2020deep} is a semantic segmentation-based searching and grasping approach without object generalization functionality. 
2) \textbf{IOSG-Hard-Thresh} is a variant of our approach where we only preserve the similarity projection map $h_t$ with values over 0.5. 3) \textbf{IOSG-Normalized-Thresh} sets the highest mask similarity score of the matched object to be 0.99, and the rest of the object score to be 0.01. 4) \textbf{IOSG-Prob-Map} uses the original similarity projection map $h_t$ to train the DQN. However, instead of using the action classifier $f_a$ from Fig. \ref{coordinator::fig}, \textbf{IOSG-Prob-Map} applies the same action classifier in \textbf{GTI} \cite{yang2020deep}. 5) \textbf{IOSG-Vector-Feature} uses the original $h_t$ in DQN. The coordinator classifier takes as input both the domain knowledge feature vector and the visual feature embedding from the DenseNet121 of the DQN. The baselines \textbf{IOSG-Hard-Thresh}, \textbf{IOSG-Normalized-Thresh}, and \textbf{IOSG-Prob-Map} use the same action classifier as \textbf{GTI}.

We evaluate our approach on the same 8 coordination test cases from \textbf{GTI} \cite{yang2020deep}. The maximum motion number for each test case is 5. We execute each test case 50 times and show the performance results in Fig. \ref{coordinator::performance}. Except \textbf{GTI} that uses the pre-trained semantic segmentation model, all other methods use SaG segmentation model \cite{yu2022self} with our target similarity network. 

Our approach, \textbf{Coordinator-Ours}, achieves 96.0\% task success rate with an average of 2.9 motions, which outperforms the other approaches by a large margin. By only changing the target object mask to the target similarity probability map, the system is not able to achieve the same level of coordination subtask performance comparing \textbf{GTI} with \textbf{IOSG-Prob-Map}. Moreover, the comparison among \textbf{IOSG-Hard-Thresh}, \textbf{IOSG-Normalized-Thresh}, and \textbf{IOSG-Prob-Map} shows that the small modifications on the similarity projection map $S$ will not improve the model performance. The comparison between \textbf{IOSG-Prob-Map} and \textbf{Coordinator-Ours} validates the choice of combining the cropped depth heightmap and the domain knowledge features for action type prediction in that the task success rate increases by $16\%$ and it requires 0.7 fewer number of motions. Although \textbf{IOSG-Vector-Feature} uses additional visual feature embedding for the coordinator classifier, the visual feature embedding evolves constantly during training so that the coordinator cannot learn meaningful action type prediction. Thus, the overall performance is lower than our approach. 

\begin{figure}[t]
    \centering
    \begin{subfigure}{0.238\linewidth}
        \includegraphics[width=\textwidth]{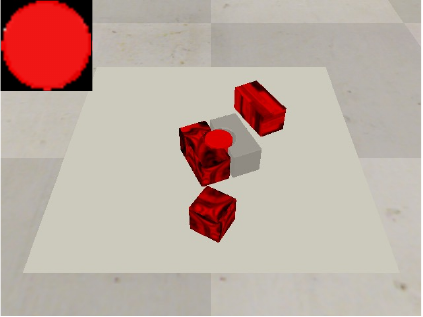}%
        \vspace{-5pt}
        \caption*{case 1}
    \end{subfigure}
    \begin{subfigure}{0.238\linewidth}
        \includegraphics[width=\textwidth]{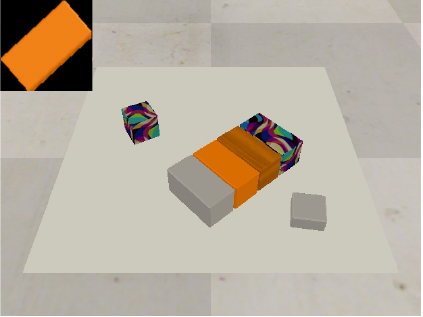}%
        \vspace{-5pt}
        \caption*{case 2}
    \end{subfigure}
    \begin{subfigure}{0.238\linewidth}
        \includegraphics[width=\textwidth]{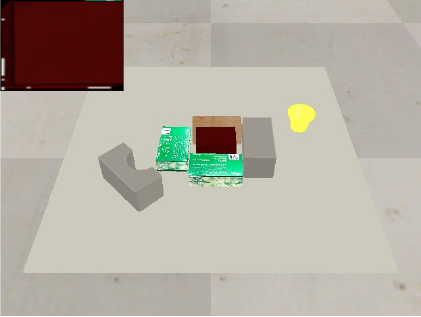}%
        \vspace{-5pt}
        \caption*{case 3}
    \end{subfigure}
    \begin{subfigure}{0.238\linewidth}
        \includegraphics[width=\textwidth]{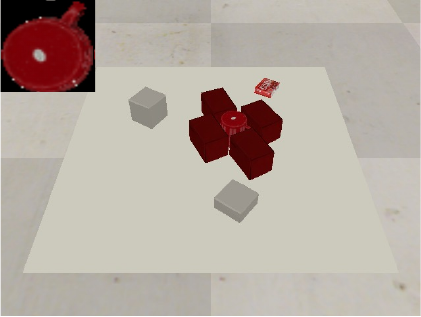}%
        \vspace{-5pt}
        \caption*{case 4}
    \end{subfigure}
    \vspace{-6mm}
    \caption{\textbf{Confusing cases.} The target object is shown on the upper left corner of each image. We apply domain randomization on some non-target objects and use YCB textured objects to make the test scenes more challenging.}
    \vspace{-5mm}
    \label{fig::confusion}
\end{figure}

\begin{figure}[h]
\includegraphics[width=\linewidth]{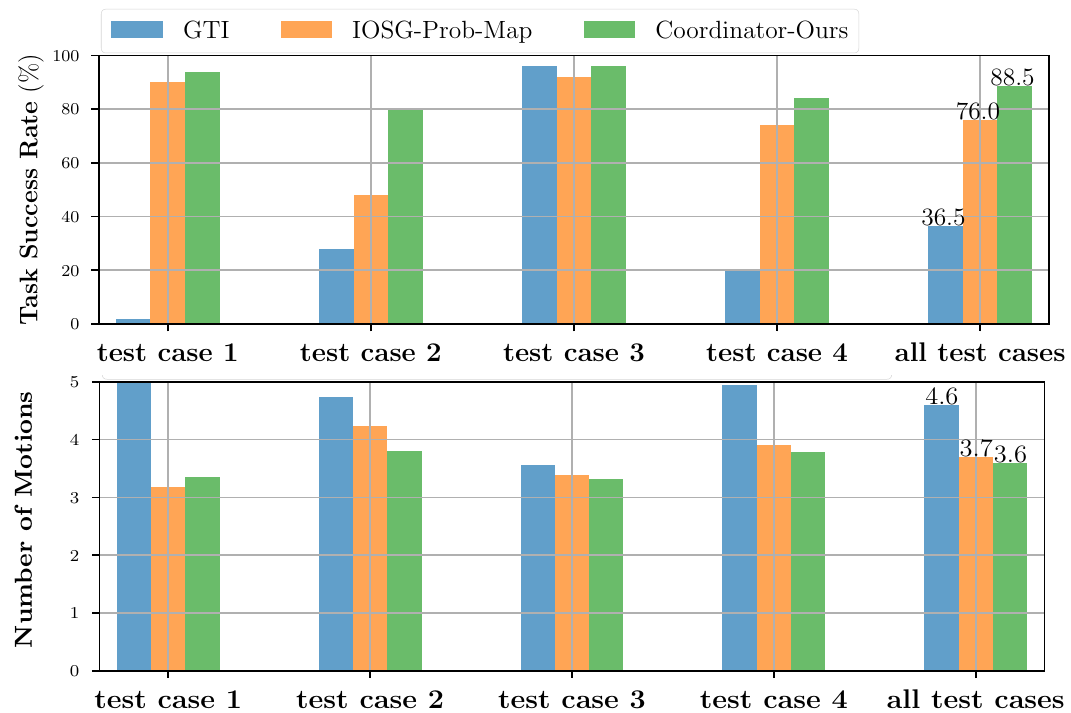}
\caption{\textbf{Performance in confusing cases}. We show the task success rate and the average number of motions in 4 confusing test cases with 3 different methods. Our proposed model achieves 88.5\% success rate with 3.6 motions.}
\vspace{-2mm}
\label{confusion::performance}
\end{figure}

\textbf{Confusing Cases:} As shown in Fig. \ref{fig::confusion}, we evaluate \textbf{GTI}, \textbf{IOSG-Prob-Map}, and \textbf{Coordinator-Ours} with additional 4 challenging testing cases, which we refer as confusing cases. In the test case $1$ and $2$, we apply domain randomization \cite{tobin2017domain, yang2021attribute} to non-target objects; while in test case $3$ and $4$ we use YCB \cite{ycb} objects with complex textures to create cluttered scenes. The maximum motion number limit is still 5. We run each test case 50 times and report the performance in Fig. \ref{confusion::performance}. Since \textbf{GTI} requires accurate target object masks, these test cases, where the objects have similar colors or unseen textures, are more challenging. \textbf{GTI} with the pre-trained Light-Weight RefineNet \cite{BMVC18Nekrasov} cannot distinguish the target object from the confusing objects, resulting in undesirable performance, especially in the confusing case $1$. In contrast, \textbf{IOSG-Prob-Map} and \textbf{Coordinator-Ours} utilize target similarity maps with instance segmentation models. Therefore, the performance of the two baselines is much higher than \textbf{GTI}. Specifically, \textbf{Coordinator-Ours} outperforms \textbf{GTI} by $52.0\%$ in task success rate and requires 1.0 less average motion number. The comparison of \textbf{IOSG-Prob-Map} and \textbf{Coordinator-Ours} also shows the effectiveness and efficiency of the coordinator policy $\pi_c$.

\begin{figure}[t]
    \centering
    \begin{subfigure}{0.238\linewidth}
        \includegraphics[width=\textwidth]{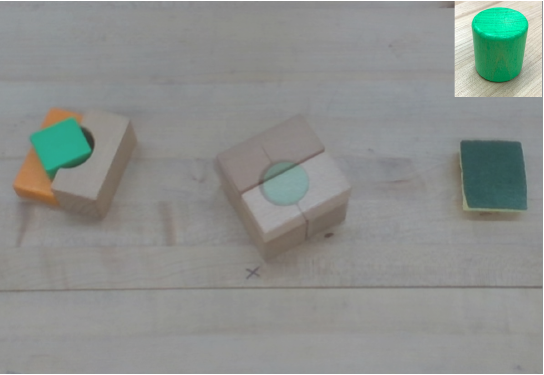}%
        \vspace{-5pt}
        \caption*{green cylinder}
    \end{subfigure}
    \begin{subfigure}{0.238\linewidth}
        \includegraphics[width=\textwidth]{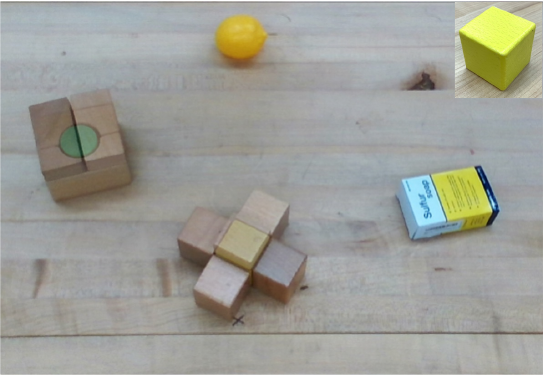}%
        \vspace{-5pt}
        \caption*{yellow cube}
    \end{subfigure}
    \begin{subfigure}{0.238\linewidth}
        \includegraphics[width=\textwidth]{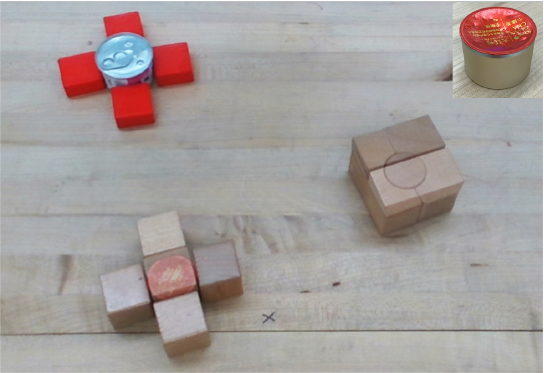}%
        \vspace{-5pt}
        \caption*{tea tin}
    \end{subfigure}
    \begin{subfigure}{0.238\linewidth}
        \includegraphics[width=\textwidth]{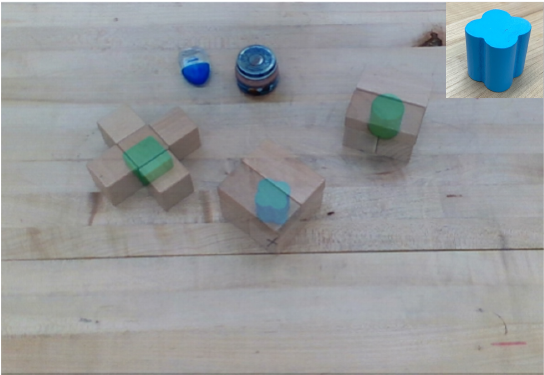}%
        \vspace{-5pt}
        \caption*{blue block}
    \end{subfigure}
    \vspace{-5mm}
    \caption{\textbf{Test cases on real robot.} The \rev{only} target object is shown on the upper right corner of each image. There are multiple confusing objects in each test case.}
    \vspace{-3mm}
    \label{fig::real}
\end{figure}

\subsection{Real Robot Experiments}

We evaluate the full target object searching and grasping task with a Franka Emika Panda robot. We directly use the policies trained in simulation without real robot fine-tuning. 
We compare our method with \textbf{GTI} \cite{yang2020deep} and \textbf{IOSG-Prob-Map}. \textbf{IOSG-Prob-Map} uses the same explorer as our approach and applies \textbf{IOSG-Prob-Map} coordinator as in Section \ref{Cooordination analysis}. 

Four challenging test cases are shown in Fig. \ref{fig::real}. \rev{In each testing case, there is only one target object with multiple confusing objects. A test trial is considered successful only if the appointed target object is grasped.}

We run each test case 10 times, and the maximum number of motions for each run is 15. The task success rate and the average number of motions are shown in Table \ref{tab3}. Our \textbf{IOSG} approach achieves a task success rate of 85.0\% with the best motion efficiency among those. The results show that our model is able to generalize to new environments and novel objects. The comparison between \textbf{IOSG} and \textbf{IOSG-Prob-Map} shows that the additional visual feature input to the coordinator classifier improves the push and grasp synergy. \textbf{GTI} is incapable of distinguishing the target object from the confusing objects. As long as there are similar objects in the scene, \textbf{GTI} tends to prioritize grasping the confusing object.

\begin{table}[h]
\caption{Real Robot Results on Searching and Grasping}
\vspace{-3mm}
\label{tab3}
\begin{center}
\begin{tabular}{ccc}
\toprule
  & Task Success Rate (\%) & Average Motions \\
\midrule
IOSG & \textbf{85.0} & \textbf{6.8}  \\
IOSG-Prob-Map & 67.5 & 9.8 \\
GTI & 35.0 & 12.6 \\
\bottomrule
\end{tabular}
\vspace{-6mm}
\end{center}
\end{table}

\section{Conclusion}
We presented an Image-driven Object Searching and Grasping (IOSG) approach trained in deep Q-learning. We used an instance segmentation model with the target similarity network to reason about the target matching given a reference image to improve the model generalization ability. We designed proper reward functions and trained the exploration policy to search the target object. To further enhance the model in target-oriented push-and-grasp, we utilized the depth heightmap information to reason about the action type generated from the coordinator policy and achieved higher coordination subtask success rate. Our approach showed a $94.5\%$ and $96.0\%$ task success rate in exploration subtask and coordination subtask in simulation and an $85.0\%$ in real robot experiments.






{
\bibliography{IEEEexample,IEEEabrv}
\bibliographystyle{IEEEtran}
}

\end{document}